\pdfoutput=1

\documentclass[11pt]{article}

\usepackage{acl}
\usepackage[utf8]{inputenc}
\usepackage{tabularx}
\usepackage[T1]{fontenc}
\usepackage{subcaption}  
\usepackage{float}
\usepackage{times}
\usepackage{latexsym}
\usepackage{tablefootnote}
\usepackage{graphicx}
\usepackage{booktabs}
\usepackage{color,soul}

\usepackage[utf8]{inputenc}
\newcommand{\ignore}[1]{}
\newcommand{\squishlist}{
 \begin{list}{$\bullet$}
  { \setlength{\itemsep}{0pt}
     \setlength{\parsep}{2pt}
     \setlength{\topsep}{2pt}
     \setlength{\partopsep}{0pt}
     \setlength{\leftmargin}{1em}
     \setlength{\labelwidth}{1em}
     \setlength{\labelsep}{0.4em} } }

\newcommand{\squishend}{
  \end{list}  }

          \hypersetup{
           breaklinks=true,   
           colorlinks=true,   
           pdfusetitle=true,  
        }

\usepackage{microtype}

\usepackage{inconsolata}
\usepackage[compact]{titlesec}
\usepackage{url}
\usepackage{cleveref}

\title{Classist Tools: Social Class Correlates with Performance in NLP}

\author{Amanda Cercas Curry\\
  MilaNLP \\
  Bocconi University \\
  \texttt{amanda.cercas@unibocconi.it}
  \\\And
  Giuseppe Attanasio \\
  Instituto de Telecomunicações \\ 
  Lisbon, Portugal \\
  \texttt{giuseppeattanasio6@gmail.com} \\\AND
  Zeerak Talat \\
  Mohamed Bin Zayed\\ University of Artificial Intelligence \\
  \texttt{z@zeerak.org} \\\And
  Dirk Hovy\\\
  MilaNLP \\
  Bocconi University \\
  \texttt{dirk.hovy@unibocconi.it}
  }

\begin{document}
\maketitle
\begin{abstract}
Since the foundational work of William Labov on the social stratification of language \citep{labov1964social}, linguistics has made concentrated efforts to explore the links between socio-demographic characteristics and language production and perception.
But while there is strong evidence for socio-demographic characteristics in language, they are infrequently used in Natural Language Processing (NLP). Age and gender are somewhat well represented, but Labov's original target, socioeconomic status, is noticeably absent. And yet it matters. We show empirically that NLP disadvantages less-privileged socioeconomic groups. 
We annotate a corpus of 95K utterances from movies with social class, ethnicity and geographical language variety and measure the performance of NLP systems on three tasks: language modelling, automatic speech recognition, and grammar error correction. 
We find significant performance disparities that can be attributed to socioeconomic status as well as ethnicity and geographical differences. 
With NLP technologies becoming ever more ubiquitous and quotidian, 
they must accommodate all language varieties to avoid disadvantaging already marginalised groups. 
We argue for the inclusion of socioeconomic class in future language technologies.
\end{abstract}

\section{Introduction}
Class or socioeconomic status is an important part of identity formation \cite{rickford1986need,bucholtz2005identity,eckert2012three}. 
Certain accents, phrases, and expressions, particularly in English, indicate upper, middle, or lower-class or other references to socioeconomic status.
This relationship was first examined by \newcite{labov1964social}. He discovered that New Yorkers with greater socioeconomic status pronounce the /R/ sound after vowels, whereas individuals with lower socioeconomic position drop it. 
He quantified this observation by asking employees in various department shops (a proxy for socioeconomic rank) for things found on the \textit{fourth floor}. Then he kept track of how many Rs were dropped in those two words. 
He discovered a clear anti-correlation between the store's (and presumably the speakers') socioeconomic standing and the quantity of dropped Rs: the higher the status, the fewer dropped Rs.

Since \citeposs{labov1964social} analysis of social stratification and language, linguistics has made concerted efforts to understand how different socio-demographic factors influence language production and perception, and how speakers use them to create identity \cite{eckert2012three}; using both naturally produced language and language from media like TV shows and films \cite{stamou2014literature} to study sociolects.
Despite a considerable body of evidence demonstrating links between language and demographic characteristics (the ``first wave'' of socio-linguistic variation studies), comparatively few socio-demographic factors have been studied in the context of natural language processing (NLP) technology. Given NLP's role as a product and a tool for understanding other phenomena, it must represent all language varieties. 

Existing research on socio-demographic characteristics has primarily concentrated on the signaling effect of certain linguistic variables like age, ethnicity, regional origin, and gender \cite{johannsen-etal-2015-cross}. 
Most of these publications focus on any bias toward the particular variable they study.
However, NLP 
hardly interacts with the second and third waves of sociolinguistics, i.e., how variation 1) shapes local identity and 2) drives language development. To narrow this gap, we use NLP to focus on socioeconomic position.

Using a dataset of 95K utterances from English-language television episodes and movies, we empirically investigate performance as a function of socioeconomic class. We study its effect on vocabulary, automated speech recognition (ASR), language modelling, and grammatical error correction. In all applications, we see strong correlations. 
Movie characters are usually typecast to present as a certain class, providing a representative sample without the privacy issues associated with regular subjects. The authenticity of these representations is backed by work in sociolinguistics such as \newcite{mchoul1987initial} and \newcite{quaglio2008television}.

We discover that the performance of NLP tools is associated with socioeconomic class, geographic variety (US vs.\ UK), and race across all tasks.
Our findings highlight an important lack of flexibility of NLP tools, and a more fundamental issue: What does it mean if NLP technologies only reliably work for a limited segment of society?

\paragraph{Contributions}
\begin{itemize}
    
    \item We construct a corpus of 95K utterances from television shows and movie scripts, coded for socioeconomic status, geography, and race.
    \item We empirically show that socioeconomic status measurably impacts the performance of NLP systems across four measures.
    \end{itemize}

\section{Social Class and Its Impact on Language}

Social stratification is the grouping of people based on their socioeconomic status (SES), which is determined by factors such as income, education, wealth, and other characteristics. Different groups are distinguished in terms of power and prestige. 
There are various social stratification systems, such as the Indian caste system, indigenous American clans or tribes, and the Western hierarchical class system. 

The exact number of social strata is unknown and is likely to vary by country and culture.
However, at least three strata are commonly used to refer to different groups in a society: upper, middle, and lower-class people. 
Other systems distinguish between blue collar and white collar jobs.  
Recently, the Great British Class Survey~\citep[GBCS,][]{savage2013new} has taken an empirical approach to understanding the different social strata, and they propose a seven-level system for the United Kingdom. 
Their stratification is based on economic, social, and cultural capital: elite, established middle-class, technical middle-class, new affluent workers, traditional working class, emergent service workers, and precariat. 
They derive these classes from a survey conducted over British citizens that received more than $160$K responses.

Social class influences people's daily lives by granting or limiting access to resources. 
Beyond power and prestige, social stratification has a significant impact on people. For example, lower socioeconomic status has been linked to poorer health outcomes and higher mortality.~\cite{Saydah_Imperatore_Beckles_2013}. 

SES also affects language use from the very early stages of development.
\newcite{bernstein1960language} posits that language takes on a different role in middle- and working-class families, where middle-class parents encourage language learning to describe more abstract thinking. 
In working-class families, parents are \emph{limited} to more concrete and descriptive concepts. 
Parents from lower SES tend to interact less with their children, with fewer open-ended questions than parents from higher SES, which shapes language development~\cite{clark2015first}. 
While \newcite{usategui1992sociolinguistica} show that education reduces this language gap, education has traditionally been one of the most important factors in determining social class and potential for upward social mobility. 

Given the well-documented effects of socioeconomic status on language development and use, it stands to reason that social class should be carefully considered as a variable in NLP.

\section{Dataset}
Here, we want to assess the impact a user's social class has on the performance of NLP systems. 
To the best of our knowledge, no datasets are available that would suit our purposes: a large enough sample that is (ethically) annotated with the speakers' socieconomic background. We thus collect our own. 
We identify an initial sample of English language television series by selecting the $250$ highest-rated series on the International Movie Database (IMDB).\footnote{\href{https://www.imdb.com/chart/toptv}{IMDB top 250 TV shows.}}
We annotate each show according to attributes,\footnote{Note that we are using \emph{ascribed} characteristics rather than \emph{self-reported} since we are dealing with fictional characters, however these two are found to be highly correlated based on linguistic features \cite{weirich2018gender}} i.e., class, race, gender, dialect, and geography of the main characters, as well as the genre and time period of the show. 
Three annotators familiar with the shows agree on the labels unanimously. 
We only consider shows that are scripted to ensure that the language production of each character is controlled by the script-writers and actors to produce a particular vernacular and to avoid collecting highly sensitive information about private individuals.
We therefore exclude cartoons, documentaries.
We further exclude science fiction and historical fiction, as these are likely to operate without strict adherence to existing social structures and language variation.
After our filtering and selection, 63 shows remain in consideration.
Of those, 34 are predominantly about white men, and 29 feature a mixture of white women and men. 
Only three predominantly feature female characters, and none black or lower-class female characters. 
Based on this initial sample, we identified additional shows to augment gaps (e.g., the lack of Black stories) in class and dialect.
Because we want to measure the effect of class separately from gender, race and regional accent, we select the TV shows from the list representing each category.


In our final selection of movies and television shows, we include a total of 19 shows that are based in the United States of America and the United Kingdom.
To ensure racial diversity, we emphasise movies and shows that predominantly represent Black characters in addition to movies and shows that represent white characters.\footnote{At least one of the authors is familiar with every show and movie. We annotated them for race and class based on the setup and main characters.}
We include the U.S.-based shows \textit{The Wire} and \textit{When They See Us} (predominantly lower-class Black characters), \textit{Fargo} (middle-class Black characters), \textit{The Fresh Prince of Bel-Air} (upper-class Black characters).
For the shows' centre white characters, we include \textit{Trailer Park Boys} (lower-class white), \textit{The Sopranos} (lower-class white), \textit{Breaking Bad} (middle-class white), and \textit{Arrested Development} (upper-class white).
The list of shows above, however, predominantly feature male characters. For this reason, we also include shows whose characters represent a wider array of genders.
The shows that we include are \textit{Pose} (lower-class Black and Latinx), \textit{Big Little Lies} (middle-class white), and \textit{Sex and the City} (upper middle-class white).
Finally, to compare with other regional varieties of English, we include movies and shows from different regions in the United Kingdom. The movies and shows that we include are \textit{Train Spotting} and \textit{T2} (lower-class white Scottish, predominantly male characters), \textit{Derry Girls} (lower-class white Irish, primarily female), \textit{Downton Abbey} (English), \textit{The IT Crowd} (white, middle-class) and \textit{The Crown} (upper-class white English). Full details in Table \ref{tab:media_demographics} in the Appendix.

We collect the first season for each of the shows on our list, except for Fargo and The Wire.
For Fargo, we collect the fourth season, which includes Black characters.
For The Wire, we collect seasons 1 and 3, which predominantly feature Black characters but have a significant minority representation of white characters.

We deliberately chose this approach to data selection from ``artificial sources'' since collecting class information is highly sensitive and can introduce privacy risks. 
Fictional characters, however, are often scripted into particular roles with explicit SES.
A drawback of this methodology is the question of authenticity  - the extent to which actors are representative of particular sociolects. Actors reportedly undergo significant training to learn different lects.\footnote{See for example, \url{https://www.newyorker.com/magazine/2009/11/09/talk-this-way}} By selecting highly rated shows, we hope to capture good, realistic performances. Moreover, TV show and film data is increasingly used in sociolinguistics \cite{stamou2014literature} and \newcite{quaglio2008television} found important similarities between real and TV show dialogue. We therefore have reason to believe our dataset provides a fair representation of real speakers.\looseness=-1

\ignore{
\begin{itemize}
    \item Class is a sensitive attribute and collecting natural content can therefore introduce privacy risks
    \item reducing risk by selecting scripted tv-shows where characters are written to express their demographic belonging
    \item 
\end{itemize}
}

Our dataset contains 95K utterances, text and speech, from 19 TV shows and movies (see \Cref{tab:statistics} for dataset statistics). 

\begin{table}[!t]
\resizebox{\columnwidth}{!}{
\begin{tabular}{llrr}
\toprule
Geography & Class  &  Episodes &  Utterances \\
\midrule

USA & Low &       13 &  20119 \\
    & Low, middle &       13 &  22947 \\
    & Middle &        7 &   4140 \\
    & Middle, Upper &       25 &  13221 \\
    & Upper &       15 &   8561 \\

EN & Middle &        6 &   2271 \\
    & Upper &        3 &   1515 \\
    & Upper, Low &        7 &   5214 \\
NE & Low &        7 &   3929 \\
    & Middle &        2 &   1238 \\
\bottomrule
\end{tabular}
}
\caption{Summary of statistics by geographic variety and class. EN: refers to England, NE to non-English U.K.}
\label{tab:statistics}
\vspace{-1.5em}
\end{table}

\subsection{Lexical Analysis}

\newcite{bernstein1960language} showed that children from working-class families had significantly smaller vocabularies even when general IQ was controlled for, and \newcite{flekova-etal-2016-exploring} showed that there is significant lexical and stylistic variation between different social strata, with lexical features being stronger predictors of income than even age. 

Thus, we first assess whether our dataset has sufficient linguistic cues to capture lexical differences in language use. 
We generally follow the methodology set out in \newcite{flekova-etal-2016-exploring}. However, we model class as a categorical value. We calculate the following features:

\paragraph{Surface:} Turn length, mean word length per turn, ratio of words longer than five letters, type-token ratio. \newcite{flekova-etal-2016-exploring} also measure the use of emojis, but this is not relevant to our analysis. 

\paragraph{Readability Metrics:} Readability metrics aim to measure the complexity of a text generally based on the number of syllables per word and the number of words per sentence. Using Textstat,\footnote{\url{https://github.com/textstat/textstat}} we calculate the Automatic Readability Index (ARI)~\cite{senter1967automated}, the Flesch-Kincaid Grade Level (FKG)~\cite{kincaid1975derivation}, the Coleman-Liau Index (CLI)~\cite{coleman1975computer}, the Flesch Reading Ease (FRE)~\cite{flesch1948new}, the LIX Index (LIX)~\cite{anderson1983lix}, and the Gunning-Fog Index (FOG)~\cite{gunning1952technique}. We normalise the text before calculating these metrics by lower-casing and removing punctuating except for periods since these are used by CLI. Note that these were designed to evaluate long-form text.

\paragraph{Syntax:} Texts with more nouns and articles as opposed to pronouns and adverbs are considered more formal. We measure the ratio of each POS per turn using Stanza~\cite{qi-etal-2020-stanza}.

\paragraph{Style:} Finally, to capture some notion of speaker style, we measure the number of abstract words\footnote{\url{https://onlymyenglish.com/list-of-abstract-nouns/}}
, the ratio of hapax legomena (HL), and the number of named entities (NER). 

\begin{table}[!t]
\resizebox{\columnwidth}{!}{%
\begin{tabular}{lrrrr}
\toprule
{} &  Class &  Gender &   Race &  Geography \\
\midrule
Length    & -0.033 & -0.034 & -0.025 &      0.052 \\
Chars & \textbf{-0.437} &  \textbf{0.243} &  \textbf{0.203} &      \textbf{0.155} \\
>5       &  0.073 & -0.024 &  0.017 &     -0.022 \\
TTR            & -0.019 & -0.047 & -0.046 &     -0.085 \\ 
\midrule
FRE            & -0.096 &  0.033 &  0.032 &      0.074 \\
ARI            & -0.214 &  0.068 &  0.059 &      0.110 \\
CLI            &  \textbf{0.467} & \textbf{-0.213} & \textbf{-0.201} &     \textbf{-0.137} \\
FKG            &  0.007 & -0.035 & -0.029 &     -0.005 \\
FOG            &  0.137 & -0.031 & -0.040 &     -0.025 \\
LIX            &  0.106 &  0.013 & -0.007 &     -0.065 \\ 
\midrule
NOUN           & \textbf{-0.181} &  0.076 & -0.035 &      0.163 \\
VERB           & -0.059 &  0.039 & -0.002 &      \textbf{0.082} \\
PROPN          & -0.017 & -0.064 & -0.099 &     -0.067 \\
ADV            & -0.066 & -0.068 & -0.103 &     -0.037 \\
ADJ            & -0.051 &  0.044 &  0.021 &      0.076 \\
DET            &  0.037 &  0.059 &  0.024 &      0.026 \\
INTJ           &  0.121 & \textbf{-0.079} & \textbf{-0.093} &     -0.044 \\
PRON           & -0.004 &  0.014 &  0.008 &      0.020 \\
NER            &  0.078 & -0.001 &  0.035 &     -0.010 \\ 
\midrule
abst           &  \textbf{0.052} & -0.034 & -0.026 &     -0.034 \\
HL             & -0.034 &  \textbf{0.109} &  \textbf{0.142} &      \textbf{0.066} \\
\bottomrule
\end{tabular}
}
\caption{Coefficients for the logistic regression model. For each class and type of feature, the strongest coefficient is in bold face.}
\label{tab:log_coefficients}
\end{table}

\begin{table}
\centering
    \begin{tabular}{lcc} \toprule
    & Val & Test \\ \midrule
   Lexical  &  & 0.290\\
    TF-IDF & \textbf{0.528} & \textbf{0.524} \\  
    TF-IDF + Lexical & 0.297 & 0.290 \\
    Sentence Embed & 0.457 & 0.454 \\ \bottomrule
\end{tabular}
\caption{F1 Macro for different representations on social class prediction. Best results in bold.}
\label{tab:logreg_results}
\end{table}

We run a multi-class logistic regression to understand whether the features are correlated to class. 
After normalisation, we run a logistic regression with social class as the predicted class and the features above as the predictive features. 
In contrast with \newcite{flekova-etal-2016-exploring}, we find  our set of features only mildly predictive in this dataset. Although CLI and the number of characters have strong coefficients, the overall performance is very low. Our best model achieves a macro F1 of 0.16, only slightly better than a most-frequent label baseline (macro-F1 of 0.07). 
In addition, we run the same regression for the other sociodemographic aspects: gender, race, and geographical origin. Overall, we find these less predictive with F1 scores similar to the baseline, suggesting the features better capture differences caused by SES than other sociodemographic factors. Table \ref{tab:log_coefficients} shows the coefficients for each sociodemographic factor. 

Lexical features do not capture strong differences in SES indicating one of two things: the features are better suited for written text (for example, many of them capture formality), and/or our dataset does not accurately capture linguistic differences. To assess whether linguistic cues in our dataset might help us differentiate between speakers of SES which the previous study did not capture, we experiment with alternative text encodings such as TF-IDF, and sentence embeddings \cite{reimers-gurevych-2019-sentence}.\footnote{We use the standard scikit-learn's \href{https://scikit-learn.org/stable/modules/generated/sklearn.feature_extraction.text.TfidfVectorizer.html}{TfidfVectorizer} collecting uni- and bi-grams for TF-IDF, and \url{https://huggingface.co/sentence-transformers/all-mpnet-base-v2}.}
The results are shown in Table \ref{tab:logreg_results}. Logistic regression based on TF-IDF achieves an F1 of 0.52 (x1.8 lexical features). 
This finding suggests there are linguistic differences, but the metrics do not capture them. 

\begin{figure*}[!t]
  \begin{subfigure}{\columnwidth}
    \includegraphics[width=\linewidth]{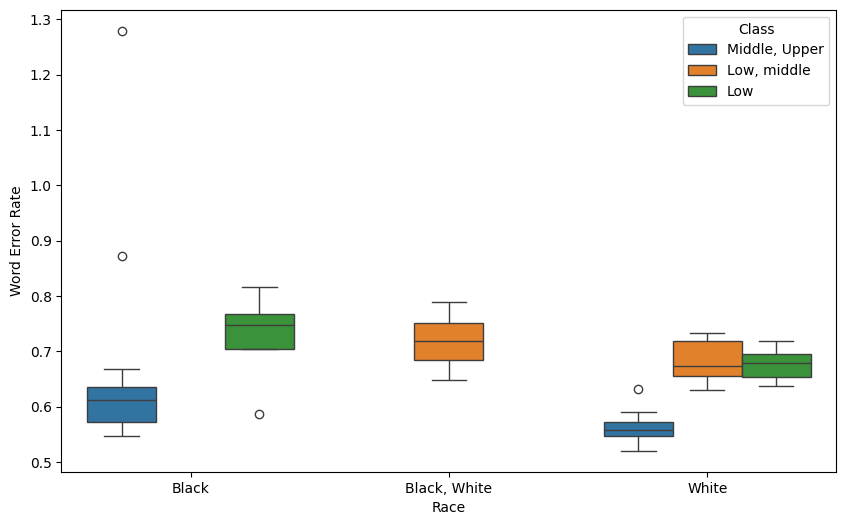}
    \caption{Wav2Vec2}
    \label{fig:sub1}
  \end{subfigure}%
  \begin{subfigure}{\columnwidth}
    \includegraphics[width=\linewidth]{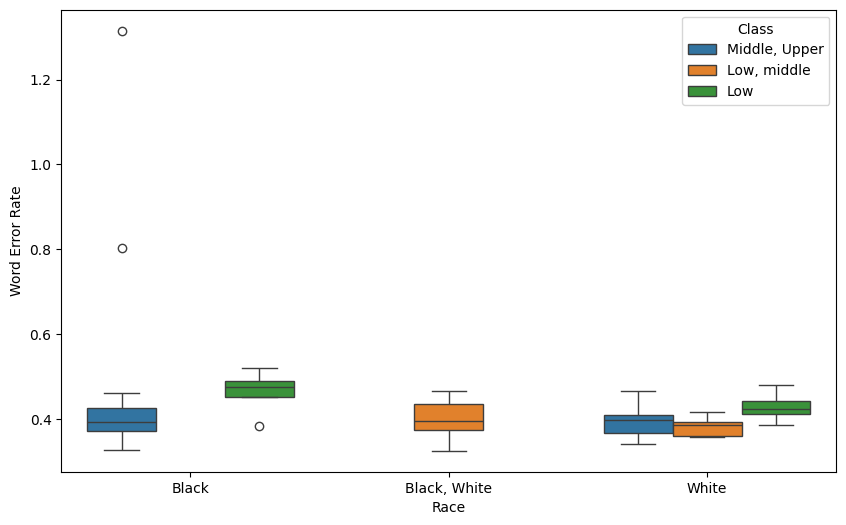}
    \caption{Whisper}
    \label{fig:sub2}
  \end{subfigure}
  \caption{Word error rate for the different models, grouped by the speakers' SES and race, for US-based TV shows. The race in the chart refers to that of the majority of characters in the show.}
  \label{fig:asr_race_class}
\end{figure*}

\section{Social Class and Speech}

Beyond written language, our dataset affords testing whether speech models understand and process accent unevenly. 
Therefore, we test disparities in ASR word error rates across social classes. 
We calculate the WER for all shows and movies to measure how well ASR technologies capture different variants of English in our sample. 

\paragraph{Models:} We use Wav2Vec2 \cite{baevski2020wav2vec}, XLS-R \cite{babu2021xls}, and Whisper-medium \cite{radford2023robust} as implemented in \texttt{transformers} \cite{wolf-etal-2020-transformers}.

\paragraph{Processing Pipeline:}
We use movies and TV shows which are available online.  
We first collect the audio and subtitles from OpenSubtitles\footnote{\url{https://www.opensubtitles.org/}} and use them as gold reference.\footnote{Here, we manually checked a random sample to verify the quality. 
We henceforth assume that the subtitles are an accurate transcription of what is being said.}
Then, we used one ASR model to transcribe the audio.

The timestamps from the model-generated transcription and the subtitles often do not match, and neither source provides information about which character is speaking.
We therefore group all utterances from a single episode or movie together into one long string and calculate the WER using Jiwer~\cite{Morris_WER_2004}. 
Ideally, we would like to separate by character to get a more fine-grained analysis. However, this approach would require us to keep mappings from utterance to character consistent across episodes. Despite several approaches, we were not able to achieve this goal with the computational power we had. Our analysis is therefore group-based, rather than character-based.

\paragraph{Results:}
Figure \ref{fig:asr_race_class} shows the differences in word-error-rate by race and class in US TV shows. We find clear trends across models, with effects from both race and class: \textbf{lower ASR error rates are associated with higher SES and with whiteness.} These trends are stronger for the Wav2Vec2 model. 

\section{Language Modelling}
Language modelling plays an essential role in downstream applications in NLP, so it is imperative that language models can accurately and equally represent different speakers' lects. Language models are evaluated through perplexity, a measure of how expected a given sentence is: the higher the perplexity, the more unexpected. Perplexity can give us an indication of how well a model might perform in downstream tasks: \newcite{gonen-etal-2023-demystifying} show that the lower the perplexity of a prompt, the better the prompt can perform the task.
Here, we calculate perplexity as a measure of linguistic acceptability, that is, how `expectable' a particular language variant might be to a given language model. Should models display differing perplexities for different groups, they would put such groups at a disadvantage. 


\paragraph{Models:}
We experiment with two state-of-the-art base models, Llama 2 \cite{touvron2023llama} and Mistral-7B \citep{jiang2023mistral}. Moreover, we include Zephyr-7B \cite{tunstall2023zephyr}, a Mistral-7B optimized assistant model to test the effect of alignment.
\footnote{Hugging Face Hub IDs: \href{https://huggingface.co/meta-llama/Llama-2-7b}{meta-llama/Llama-2-7b}, \href{https://huggingface.co/mistralai/Mistral-7B-v0.1}{mistralai/Mistral-7B-v0.1}, and \href{https://huggingface.co/HuggingFaceH4/zephyr-7b-beta}{HuggingFaceH4/zephyr-7b-beta}}


We calculate perplexity for each sentence in our dataset after lower-casing them and removing numbers and punctuation. We exclude turns shorter than five tokens as they generally do not differentiate between classes. 

\paragraph{Results:} Table \ref{tab:perplexity} shows the mean perplexity and standard deviation for each model and each class for the entire dataset. We do not find significant differences based on geographical location (U.K. vs U.S.A) across the models. We see small differences across groups based on class alone, however, because language is affected by other socio-demographic besides class, we also consider the effect of race and geographic variety. 

\begin{figure}[!t]
    \centering
    \includegraphics[width=\columnwidth]{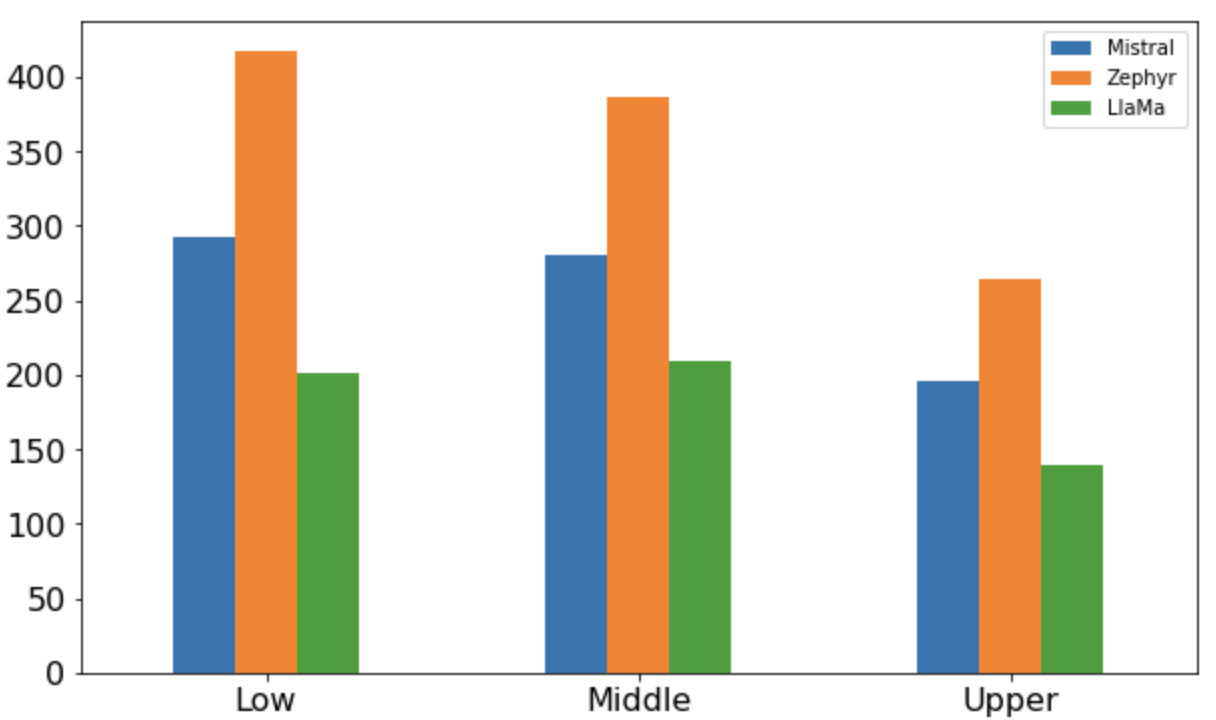}
    \caption{Mean perplexity among U.K.-based shows by model.}
    \label{fig:uk_perplexity}
\end{figure}

We calculate mean perplexity by SES in the U.K. and the U.S.A. Figure \ref{fig:uk_perplexity} shows that class and perplexity are correlated and we find similar patterns across geographical dialects: higher SES leads to lower perplexity also in U.S.A.-based shows. 

Next, we consider U.S.-based shows. The breakdown of by class and race are shown in Table \ref{tab:perplexity_by_race}. Once both factors are taken into account, a clearer picture emerges: \textbf{lower SES leads to higher perplexity.} To confirm these differences are significant, we run a one-tailed student's t-test and find significant differences between classes, particularly for white speakers for both Mistral-7B and Zephyr-7B. In the case of Mistral, lower-class and lower-middle class speakers have significantly higher perplexities than Mid-Upper ($p<0.05$) and Middle class peakers ($p<0.05$). In the case of Zephyr, the model shows higher perplexities for lower classes than it does for Middle ($p<0.05$), Mid-Upper ($p<0.01$) and Upper class speakers ($p<0.01$).

\begin{table*}[!t]
\centering
    \begin{tabular}{lrrrrrr}
    \toprule
    \multicolumn{1}{c}{} & \multicolumn{2}{c}{Mistral-7B} & \multicolumn{2}{c}{Zephyr-7B}                        & \multicolumn{2}{c}{Llama 2}                        \\ 

    {Class} &   Mean &      Std &   Mean &      Std &  Mean &    Std \\
    \midrule
Low           & 294.606 & 690.361 & 415.641 &  989.066 & 189.804 & 361.815 \\
Low, middle   & 442.756 & 911.300 & 649.462 & 1441.759 & 265.114 & 477.377 \\
Middle        & 252.729 & 536.903 & 353.667 &  822.553 & 177.900 & 398.560 \\
Middle, Upper & \textbf{241.923} & 683.345 & \textbf{332.224} &  999.137 & \textbf{164.807} & 450.446 \\
Upper         & 288.324 & 693.994 & 399.854 & 1063.942 & 190.536 & 370.958 \\
Upper, Low    & 282.815 & 575.833 & 385.273 &  876.126 & 190.696 & 358.952 \\
    \bottomrule
    \end{tabular}
    \caption{Mean perplexity and standard deviation per class and model. Best mean result per model in bold.}
    \label{tab:perplexity}
\end{table*}

\begin{table}[h]
\centering
\begin{tabular}{llr}
\toprule
             &     &    Perplexity \\
\midrule
Black & Low & 328.848 \\
             & Middle, Upper & \textbf{259.673} \\ \midrule
White & Low & 275.337 \\
             & Low, middle & 342.865 \\
             & Middle & 229.278 \\
             & Middle, Upper & \textbf{205.585} \\
             & Upper & 302.057 \\
    \bottomrule
    \end{tabular}
    \caption{Mean perplexity and standard deviation per class and race. Best results per race in bold.}
    \label{tab:perplexity_by_race}
\end{table}

\section{Grammar Correction}
Finally, we consider the grammaticality of each sentence. We suspect that grammar features correlate strongly with class, as ``correct'' grammar is typically a hallmark of signalling higher SES. 
We calculate the edit distance between the original sentence and the corrected one. Lower edit distance means higher grammaticality. 
\paragraph{Models:}
Since we are concerned with potential end-user experience, we choose to evaluate the four most downloaded models on Huggingface since they will reach the largest audience. We use the following models for grammar correction:
\begin{itemize}
    \item T5 Grammar Correction\footnote{\href{https://huggingface.co/vennify/t5-base-grammar-correction}{T5-based Grammar Correction}}: A model based on the HappyTransformer\footnote{\href{https://github.com/EricFillion/happy-transformer}{HappyTransformer}} trained on the JFLEG dataset \cite{napoles-etal-2017-jfleg}.
    \item Gramformer:\footnote{\href{https://github.com/PrithivirajDamodaran/Gramformer}{Gramformer}} A seq2seq model fine-tuned on a dataset of WikiEdits. 
    \item CoEdit-large\footnote{\href{https://huggingface.co/grammarly/coedit-large}{CoEdit}}: A flan-t5-large based model, finetuned on the CoEdit dataset \cite{raheja-etal-2023-coedit}. 
    \item Grammar-synthesis-large\footnote{\href{https://huggingface.co/pszemraj/flan-t5-large-grammar-synthesis}{Flan-T5 Correction}}: A fine-tuned version of Google's flan-t5-large for grammar correction, trained on an expanded version of the JFLEG dataset. 
\end{itemize}

\begin{table}
\resizebox{\columnwidth}{!}{
\begin{tabular}{lrrrr}
\toprule
{} & Happy  &  CoEdit &  Flan-T5 &  Gramf. \\
\midrule
Low           &           2.455 &            1.963 &            6.424 &                2.192 \\
Mid-Low   &           2.590 &            2.138 &           12.050 &                1.553 \\
Middle        &           2.068 &            1.785 &            6.117 &                1.653 \\
Mid-Upper &           2.866 &            1.944 &            7.952 &                2.091 \\
Upper         &           1.122 &            1.212 &            5.518 &                0.808 \\ \midrule
\% corrected   &           19.76 &            35.94 &            66.42 &                19.11 \\
\bottomrule
\end{tabular}
}
\caption{Mean edit distance between the subtitle reference and the grammar-corrected utterance generated by each model, and percentage of sentences with at least one correction.}
\label{tab:grammar_means}
\end{table}

\begin{figure}
    \centering
    \includegraphics[width=\columnwidth]{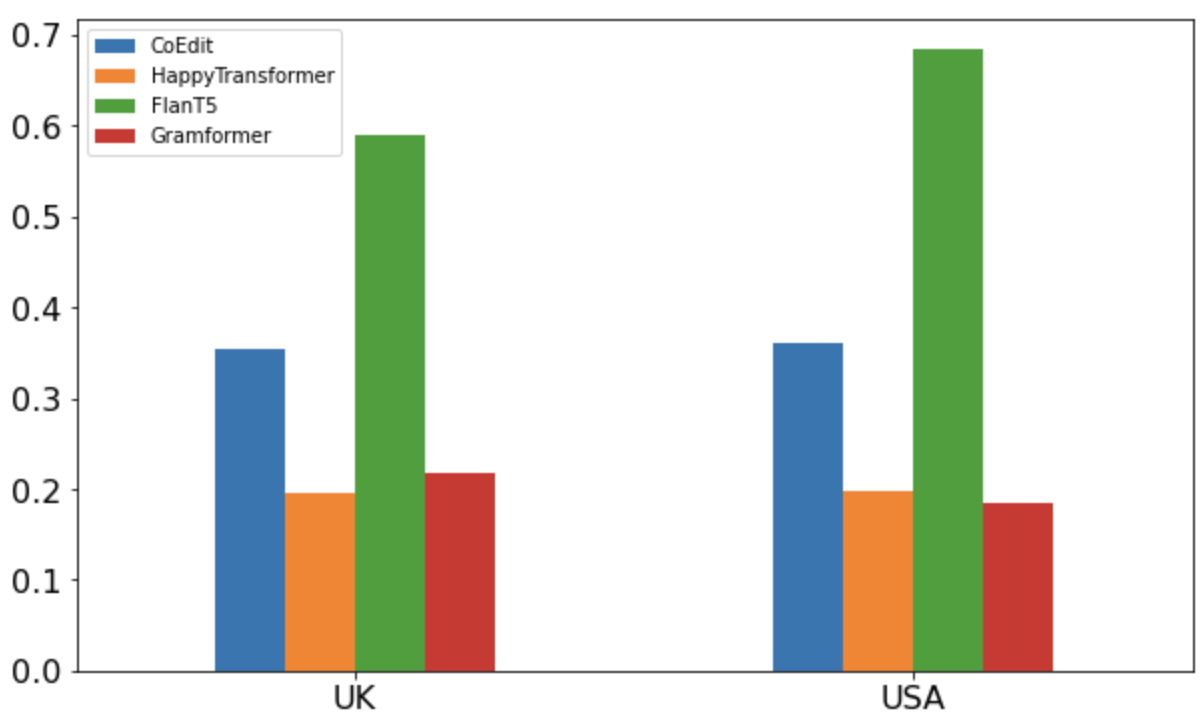}
    \caption{Proportion of utterances that are grammar-corrected per model and geographical language variety.}
    \label{fig:proportion_gram}
\end{figure}
\paragraph{Results:} Figure \ref{fig:proportion_gram} shows the percentage of utterances that each model corrected, grouped by geographical variation. The Flan-T5-based model generates significantly more corrections than the other models, followed by CoEdit-large. Overall, U.K. utterances are corrected slightly more often by the models. Table \ref{tab:grammar_means} shows the mean edit distances between the original sentence extracted from the subtitles and the models' proposed corrections.  We find little difference in grammar error correction for most models, even when controlling for geographical and racial differences. However, on further analysis, we find that models generally produce corrections for relatively few utterances.

\begin{figure}
    \centering
    \includegraphics[width=\columnwidth]{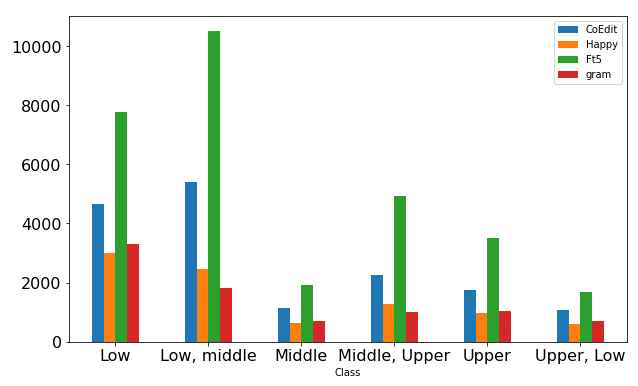}
    \caption{Distribution by class of utterances with at least one correction.}
    \label{fig:grammar_distribution}
\end{figure}

Instead, we consider whether models are more likely to produce edits for different classes. In Figure \ref{fig:grammar_distribution} we plot the count of utterances with corrections per model and per class. From this, \textbf{we can see a clear pattern where models produce corrections more frequently for those of lower SES.} 

In addition, we notice that often some of these corrections are performed on in-group slang or regional linguistic phenomena, see Table \ref{tab:grammar_correction_examples}. \\

\begin{table}[h]
    \centering
    \begin{tabular}{p{7.5cm}}
         \textbf{The Crown}  \\\hline
\textit{And what a bunch of ice-veined monsters \colorbox{pink}{my family are.}}\\
\textit{And what a bunch of ice-veined monsters \colorbox{pink}{my family is.}} \\
\vspace{0.02cm}\textbf{Pose} \\\hline
\textit{The \colorbox{pink}{tittiness} of it all.}\\
\textit{The \colorbox{pink}{quality} of it all.} \\ 
\vspace{0.02cm}\textbf{Trainspotting}  \\  \hline
So, I'm off \colorbox{pink}{the skag}.\tablefootnote{\textit{Skag} refers to heroin in Scottish slang. } \\
So, I'm off \colorbox{pink}{the grid.} \\
\vspace{0.02cm}\textbf{Trailer park Boys} \\ \hline
\textit{We \colorbox{pink}{be dope} when the new NW A come out, know what \colorbox{pink}{I'm sayin'} } \\
\textit{We \colorbox{pink}{will be happy} when the new NWA comes out, know what \colorbox{pink}{I'm saying?} }\\ 
    \end{tabular}
    \caption{Examples of grammar correction in different shows. The corrected examples are from the T5 Gramar correction model.}
    \label{tab:grammar_correction_examples}
\end{table}

\section{Discussion}
Following established work in sociolinguistics, our results support our hypothesis that NLP systems display biases regarding different language varieties beyond geographical and ethnic differences.
When it comes to evaluating the justice and fairness of NLP systems, \textbf{all} factors affecting language should be considered. 
By dint of their use, NLP systems are setting a standard of language: NLP systems are becoming more common, not only in social contexts calling for formal (i.e. standardised) language but in everyday scenarios. For example, one of the proposed use cases for the Gramformer include correcting text messages as a built-in feature for messaging apps, language models inform systems like automatic correction, and large language models like chatGPT now boast millions of users. This widespread interaction with language technologies reinforces the lect of middle class, white English speakers as the only valid English. Moreover, when we consider findings such as those by \newcite{gonen-etal-2023-demystifying}, which showed that perplexity of a model and the accuracy of the response are negatively correlated, and certain parts of the population have significantly higher mean perplexities, we risk disadvantaging those groups.

With our work, we are not encouraging more work toward social class prediction from text but to bring awareness to the NLP community about an essential factor of language use that has thus far been mostly ignored in our field. While reporting on the socioeconomic background of study participants and dataset contributors is more complicated than other factors such as age or ethnicity (as it is not a one-point measurement and it is culture-dependent), we encourage researchers to report on this to draw a more accurate picture of the language varieties NLP will capture.
Given the limitations of our study (most notably, restriction to English and group rather than individual scores, see below), we envision several possible avenues for future research in this area: (1) the findings of this study should be validated for other languages where such differences have been observed; (2) the collection of a (possibly larger) corpus of unaggregated speakers; and (3) while we find support from previous research such as \citet{flekova-etal-2016-exploring} 
for social media data, the findings should also be validated for other modalities of data such as long-form written texts.

\section{Related Work}
While there has been an uptick in the number of papers tackling gender and racial bias in NLP, work considering other under-privileged communities has lagged behind. \citet{curto2022ai} investigate bias against the poor in NLP but they do so in terms of the language that surrounds poor people, considering the associations of `poor' and `rich' in embeddings and language models. 
As a way to mitigate and document biases in NLP, \newcite{bender-friedman-2018-data} ask for the socioeconomic status of both the speakers and the annotators to be declared, however they do not suggest any standardised way to measure or report this.
\newcite{field-etal-2021-survey} 
conduct a survey focus on race but also call for more diversity in NLP in terms of the broader inclusion other underprivileged people such as those from lower socioeconomic status. To the best of our knowledge, we are the first to investigate how NLP systems fare when faced with different SES.

\section{Conclusion}
Social class plays an important role in people's identity construction, and is consequently strongly reflected in their language use. Despite its central status as a variable in sociolinguistics, there is so far little work in NLP engaging with social class and its impact. 
Our contributions are twofold: following methods from sociolinguistics, we use a data set of 95K movie transcripts that we annotate for portrayed class, race, and geographical variety (UK vs.\ US). We use this data to show that linguistic class markers greatly affect the performance of various NLP tools. 
Our findings suggest that speakers of low-prestige sociolects experience lower application performance on a range of tasks.
These results suggest that we should actively incorporate social class as a variable in NLP systems if we want to make them more equitable and fair. Our data set also provides a starting point for more explorations of social class in NLP.

\section*{Limitations}
Our study presents several limitations:
First, we are limited to the study of class, region, and race. 
TV shows and films are biased in their representations of different identities---not all groups are (equally) represented and those that are may be 
stereotyped into certain roles. 
For example, in the initial top-200 TV shows list, we did not 
find 
any TV shows about lower-class women, Latinxs, 
upper-class Non-English Britons, or non-white Britons. 
For this reason, we excluded a gender-based analysis.
Moreover, in this list, the majority of TV shows about Black Americans revolve around the drug trade.\looseness=-1

Second, characters are not necessarily written by, nor perfectly depict the group they represent and this is potentially reflected in their language. 
To ensure high quality portrayals, we sourced highly rated and realistic shows, ensuring characters were not caricatures of the groups they portrayed. 
Having said that, a sociolinguistics review found that TV show and film data is increasingly common as data for the study of sociolects, particularly with regards to race and class \cite{stamou2014literature}. 
More specifically, work in sociolinguistics has also defended the use of film data in conversation analysis \cite{mchoul1987initial, quaglio2008television}. 
Furthermore, TV shows like \textit{The Wire} have been praised for their linguistic authenticity \cite{lopez2017my}, reportedly primarily hiring Baltimore natives. 
For these reasons, we believe our study sufficiently captures the relevant linguistic phenomena. 

Despite our best efforts, we were unable to reliably separate the utterances by character across episodes. Instead of introducing more sources of potential mix-ups, we therefore estimate the metrics over all characters in a show. Although this is a limitation in terms of the accuracy of the results, we expect that character-level annotations would strengthen our findings rather than negate them. Generally speaking, each TV show focuses on a single social stratum. A single character belonging to a different stratum would add noise and make the classes less clearly defined, so we suspect our results would be more marked with character-level annotations, not less. 
We leave the character-based separation for future work. 

Finally, though our dataset is large in terms of utterances, it is limited in the number of characters. However, we present a case study hoping to motivate further research that may validate our findings.

\section*{Ethical Considerations}
Our paper is arguing for fairness in the services provided by NLP systems for people of all socioeconomic backgrounds. However, we note that by showing measurable differences in language use across groups, we run the risk of profiling speakers. In our work, we used fictional characters to avoid this, but in order to ensure fairness of service all peoples must be represented. Future work should ensure that this is done respectfully and with consent from any participants.

Furthermore, measuring socioeconomic status is not a straightforward process. The class system used in this paper is fuzzy and tied to Western social structures that are not valid across the world. Future work should focus on measuring in the most appropriate way by following established metrics and guidelines from economics and other fields, such as ~\newcite{savage2013new}.
We urge any future work to follow ethical guidelines when it comes to dataset and system development.

\bibliography{custom,anthology}

\appendix

\section{Annotations per show}
\label{sec:appendix}
Table \ref{tab:media_demographics} shows the annotated demographics for each season/film. 
\begin{table*}[]
    \centering
   \begin{tabular}{lllll} \toprule
& Class & Gender & Race & Geography \\ \midrule
Arrested Development & Upper & M, F & White & USA \\
Sex and the City & Middle, Upper & F & White & USA \\
The Fresh Prince of BelAir & Middle, Upper & M, F & Black & USA \\
Big Little Lies & Middle & F & White & USA \\
Breaking Bad & Low, middle & M & White & USA \\
The Wire (S01) & Low, middle & M & Black, white & USA \\
The Wire (S03) & Low, middle & M & Black, white & USA \\
Fargo (S04) & Low, middle & M, F & Black, white & USA \\
Trailer Park Boys & Low & M & White & USA \\
When They See Us & Low & M, F & Black & USA \\
The Sopranos & Low & M & White & USA \\
Pose & Low & F, Trans & Black,Latino & USA \\ \midrule
The Crown & Upper & M, F & White & UK \\
Downton Abbey & Upper, Low & M, F & White & UK \\
The IT Crowd & Middle & M & White & UK \\
Shetland & Middle & M, F & White & Scotland (UK) \\
T2.trainspotting & Low & M & White & Scotland (UK) \\
Trainspotting & Low & M & White & Scotland (UK) \\
Smother & Middle & M, F & White & Ireland (UK) \\
Derry Girls & Low & F & White & Ireland (NE) \\
\bottomrule
\end{tabular} 
    \caption{Annotated demographics for each movie or TV show. M denotes Men and F, women. }
    \label{tab:media_demographics}
\end{table*}

\end{document}